%% file: main.tex
\documentclass[wcp]{jmlr}

\usepackage{longtable}

\usepackage{booktabs}
\usepackage{tikz}
\usetikzlibrary{positioning,arrows.meta,calc}

\makeatletter
\let\Ginclude@graphics\@org@Ginclude@graphics 
\makeatother

\title[Boosted autoencoders]{How to boost autoencoders?}

  \author{\Name{Sai Krishna} \Email{ee17b035@smail.iitm.ac.in}\\
  \Name{Thulasi Tholeti} \Email{ee15d410@ee.iitm.ac.in}\\
  \Name{Sheetal Kalyani} \Email{skalyani@ee.iitm.ac.in}\\
  \addr  Department of Electrical Engineering,
Indian Institute of Technology Madras.
 }
\usepackage{bm}

\newcommand{\tadd}[1]{{#1}}
\newcommand{\taddRev}[1]{{#1}}

\usepackage{algorithm}
\usepackage{epstopdf}

\begin{document}

\maketitle

\begin{abstract}
\tadd{
Autoencoders are a category of neural networks with applications in numerous domains and hence, improvement of their performance is gaining substantial interest from the machine learning community. Ensemble methods, such as boosting, are often adopted to enhance the performance of regular neural networks. In this work, we discuss the challenges associated with boosting autoencoders and propose a framework to overcome them. The proposed method ensures that the advantages of boosting are realized when either output (encoded or reconstructed) is used. The usefulness of the boosted ensemble is demonstrated in two applications that widely employ autoencoders: anomaly detection and clustering. }
\end{abstract}

\section{Introduction}
Autoencoders (AEs) are a class of neural networks in which the input data is encoded to a lower dimension and then decoded to reconstruct the original input. Introduced by \cite{rumelhart1986learning} in the context of backpropagation without supervision, AEs are now being used in many applications such as dimensionality reduction, classification, generation, anomaly detection, clustering, information retrieval, etc. in varied fields like image and video processing, communication systems, recommendation systems and many more \cite{bank2020autoencoders}. Given its increasing importance, there has been a lot of focus on enhancing the performance of AEs, especially in the context of prevention of overfitting. Regularized AEs, where the network was trained with a \(\mathcal{L}_1\) norm regularized reconstruction error, was employed to induce sparsity that prevented overfitting \cite{alain2014regularized}. \cite{vincent2008extracting} proposed a De-noising AE where random noise is deliberately added to the input and \cite{hinton2012improving} recommended incorporating dropout during training. 

Boosting is a time-tested method to decrease both bias and variance. It has been established that boosting neural networks makes them less prone to overfitting (\cite{DBLP:conf/birthday/Schapire13}). Boosting autoencoders is challenging as both the hidden representation and/or the reconstructed output may be used depending on the application. We explore boosting of autoencoders and develop an ensemble architecture that is applicable across the different applications. Boosting of autoencoders has been studied in the specific context of unsupervised anomaly detection by \cite{sarvari2019unsupervised}, \cite{minhas2020semi} and \cite{chen2017outlier}. \tadd{While boosting emphasises on learning data points that are hard to classify, these methods reduce the focus on the hard-to-classify data points. This is done to ensure that the ensemble network is not trained on anomaly points.} Hence, these approaches are tailor-made to detect anomalies and cannot be applied to other applications of autoencoders.

In this work, we propose an architecture to boost autoencoders. \tadd{To the best of our knowledge, the proposition of boosting autoencoders with universal applicability has not been explored before.} We demonstrate that the boosted autoencoder attains lower reconstruction error on unseen data when compared to a single network. We also illustrate the utility of the proposed ensemble network in two different applications: anomaly detection and clustering, which use the reconstructed data and the encoded representation respectively. We compare their performances with the state-of-the-art methods and derive conclusions.

\section{Background}
\subsection{Autoencoders}
AEs are multi-layered neural networks that typically perform hierarchical and nonlinear dimensionality reduction of data to yield a compressed latent representation. An autoencoder consists of two parts: 1) encoder and 2) decoder. The encoder maps the input data to a lower dimensional space and the decoder converts the encoded output back into the input dimension. Both the encoder and the decoder are trained so that the input is reconstructed at the output while obtaining an informative, compressed representation. The schematic of a typical AE is depicted in Fig. \ref{fig:schematic}. $\bm x, \bm y \in \mathbb{R}^d$ are the input and the reconstructed output. The encoded output is denoted by \(\bm h \in \mathbb{R}^m, m<d\).

\input{autoencoder_schematic}

Based on the recent survey on AEs by \cite{bank2020autoencoders}, they are classified based on various aspects such as architecture (feed forward, convolutional AEs), regularization (sparse AEs, contractive AEs) and training methods (variational AEs). In our work, we focus on enhancing the performance of feedforward and convolutional AEs using boosting.

\subsection{Boosting}
Ensemble learning, which uses multiple decision models instead of a single one, was introduced to reduce the bias and variance in a classifier. Boosting is a method of ensemble learning where the decisions of weak learners are combined to form a strong learner. The most popular algorithm for boosting, AdaBoost was introduced by \cite{freund1997decision} for a binary classification problem. It consists of a sequence of weak learners. The AdaBoost algorithm operates by maintaining a distribution of weights ($w_i's$)over the training data; the distribution is updated such that the weights of the data points which are difficult to classify are increased. This ensures that the subsequent learners focus on the data points that are prone to error. A weighted average of the decisions by each of the encoders is computed as the output. AdaBoost is listed below as Algorithm \ref{alg:adaboost}.

\SetKwInput{KwInput}{Input}                
\SetKwInput{KwOutput}{Output}  
\SetKwInput{KwInitialization}{Initialization}  
\begin{algorithm}[H]

  \KwInput{No of Classifiers M, Training data $x_1,x_2, \cdots x_n$}
  \KwInitialization{Initialize the observation weights $w_{i}=1 / n, i=1,2, \ldots, n .$}

  \For{\(m = 1,2,\cdots M\)}
    {
        Fit a classifier $T^{(m)}(\boldsymbol{x})$ to the training data using weights $w_{i}$\\
        Compute
        $e r r^{(m)}=\sum_{i=1}^{n} w_{i} \bar{I}\left(c_{i} \neq T^{(m)}\left(\boldsymbol{x}_{i}\right)\right) / \sum_{i=1}^{n} w_{i}
        $\\
        Compute
        $\alpha^{(m)}=\log \frac{1-e r r^{(m)}}{\operatorname{err}^{(m)}}
        $\\[1PC]
        Set
        $w_{i} \leftarrow w_{i} \cdot \exp \left(\alpha^{(m)} \cdot \mathbb{I}\left(c_{i} \neq T^{(m)}\left(\boldsymbol{x}_{i}\right)\right)\right), i=1, \ldots, n
        $\\
        Re-normalize $w_{i}$
    }

  \KwOutput{$
C(\boldsymbol{x})=\arg \max _{k} \sum_{m=1}^{M} \alpha^{(m)} \cdot \mathbb{I}\left(T^{(m)}(\boldsymbol{x})=k\right)
$}
\caption{Adaboost Algorithm}\label{alg:adaboost}
\end{algorithm}

\section{Boosting Autoencoders}
It has been observed that traditional AEs, such as fully connected AEs, are weak when implemented on a high dimensional dataset; on the other hand, Deep Convolutional AEs tend to overfit towards identity, even if the model capacity is limited (\cite{steck2020autoencoders}). \taddRev{In this section, we propose boosting autoencoders as a solution to both enhancing the performance of AEs and reducing the tendency to overfit. The architecture should allow the ensemble network to cater to applications which use either the reconstructed or the encoded output. This makes the task of choosing the architecture of the ensemble very crucial. In this section, we present an architecture for an ensemble of AEs, propose an algorithm for boosting them and provide simulation results.}

\subsection{Architecture of the ensemble}
The novelty of our approach lies in the architecture of the ensemble. The challenge in designing an architecture for boosting AEs is to ensure that the dimension of the compressed representation remains unchanged in spite of using multiple networks. Consider an ensemble of $M$ AEs, boosted using the output at the decoder. In this case, for an unseen data point, the latent representations of all $M$ AEs contain information about the data. Hence, the dimension of the latent representation is scaled up by a factor of $M$. To eliminate this, we propose using an ensemble of $M$ encoders and use their average output as the latent representation. Note that the proposed ensemble consists of a single decoder. The architecture proposed ensemble is illustrated in Fig. \ref{fig:ensemble}. The algorithm for boosting AEs using the proposed architecture is described in the next section.
 \input{ensemble_schematic.tex}

\subsection{Proposed algorithm}
The proposed algorithm is inspired from AdaBoost and employs an ensemble of $M$ networks (here, encoders). A distribution is maintained over the data points that suggest which data points need greater focus from the next encoder. While AdaBoost uses classification error to assign weights to data points, the proposed algorithm uses Mean Square Error (MSE) between the input and the output of the decoder, which is also termed as the reconstruction error; this is done so as to enable the algorithm to cater to a variety of applications including, but not limited to, image classification. The proposed algorithm is listed as Algorithm \ref{alg:ensemble}.

The weights corresponding to the distribution over the data points $\bm x_i$, $i=1,\cdots n$ are denoted by $w_i$. The algorithm is initialized by assigning equal weights to all the data points. For every iteration of the algorithm, data points are sampled using these weights $w_i$. Initially, the input is passed through a single encoder-decoder pair and they are trained. The weights are then re-distributed such that the samples with high reconstruction error are more likely to be sampled at the next iteration. During the next cycle, the data is passed through the first two encoders and their average is decoded. The second encoder is trained using the reconstruction error so obtained. This process is continued until all the $M$ encoders are trained. Note that only the encoder $m$ gets trained in a given cycle $m$ even though the encoded output is an average of encoders from $1$ to $m$; the decoder, on the other hand, is being constantly trained.

\SetKwInput{KwInput}{Input}                
\SetKwInput{KwOutput}{Output}  
\SetKwInput{KwInitialization}{Initialization}  
\begin{algorithm}[h]

  \KwInput{No of Encoders M, Training data $x_1,x_2, \cdots x_n$}
  \KwInitialization{\\1) Initialize a set of Encoders $E_1, E_2... E_M$ and Decoder $D$ with weights randomly sampled from $\mathcal{N}(0,1)$ \\
        2) Initialize weights to each input in the training dataset as $w_{i}=1 / n, i=1,2, \ldots, n$.}

  \For{\(m = 1,2,\cdots M\)}
    {
        \For{ \(iter = 1,2,\cdots I\) }
        {
            Obtain a batch of $Q$ training samples distributed according to  $w_{i}$'s.\\
            Pass the chosen samples to the Encoders $E_1, E_2,..$ till $E_m$.\\
            Compute $ Avg_{i} =\frac{\sum_{j=1}^{m}E^{(m)}\left(\boldsymbol{x}_{i}\right)}{m}$ for all $\boldsymbol{x}_{i}$ and pass it to the Decoder $D$.\\
            Compute MSE loss between $x_i$ and the corresponding Decoded outputs $D(Avg_{i})$\\
            Back-Propagate MSE through the decoder $D$ and the encoder $E_m$ (Only the last Encoder)
        }
        Compute $ w_{i}=\left({x}_{i} - D\left(Avg_{i})\right)\right)^2$ for every $x_i$ and re-normalize such that $\sum_i w_i=1$.
    }
  \KwOutput{The average of all encoders is the encoded output.}
\caption{Proposed Boosted Autoencoder Algorithm}\label{alg:ensemble}
\end{algorithm}

In our algorithm, the relationship between the first encoder and decoder is no different than a single AE. However, the subsequent encoders work on tweaking/correcting the output of the first encoder, so that the overall reconstruction is better. In other words, subsequent encoders learn the slack of previous encoders (Sequential Learning). At every stage, we take average of all the encoders as our latent representation. By doing this, we are enforcing every encoder to be equally represented. Note that although the encoded output is reported as an average, the individual encoders learn vastly distinct representations. This is because each encoder learns from a different distribution on the training data based on $w_i's$. The encoded output needs to be a function of the output of all encoders; for our algorithm, we have chosen the average as the function. The algorithm may be modified to use other functions in place of average. Although the procedure is similar to Adaboost, we are not trying to boost multiple weak learners in our method. The main idea at play here is Sequential Learning.

\subsection{Simulation results}
To illustrate the utility of our proposed method, we perform experiments on two well-known image datasets- CIFAR-10 (\cite{krizhevsky2009learning}) and Fashion-MNIST (\cite{xiao2017fashion}). The reconstruction error obtained by using the proposed algorithm is compared with that of a single AE.

\subsubsection{Experiments on the CIFAR-10 Dataset} 
The CIFAR-10 dataset contains 60,000 32x32 color images (in 10 different classes) of which 40000 images were used for training, 10000 for validation and 10000 images for testing. Let \textbf{Conv2D(\textit{i,o,k,s,p})} denote a 2D convolution layer where 'i' is the number of input channels, 'o' is number of output channels, 'k' is the size of kernel, 's' is stride and 'p' is padding. The architecture of the encoder used is Conv2D(3,8,4,2,1)-Conv2D(8,16,4,2,1)-Conv2D(16,16,4,2,1). The decoder is constructed symmetrically to that of the encoder. 

The architecture also uses uses a mixture of sigmoid and ReLU activation functions to overcome the Dying ReLU problem and the vanishing gradient problem cause by sigmoid (\cite{chen2017outlier}). The Adam optimizer (\cite{kingma2014adam}) was used in training both the single and the ensemble of AEs with a learning rate of 3e-3. A single AE was trained for 50 epochs using a batch size of 50. For the proposed boosted AE, we consider $M=20$; 50 data samples are chosen for each iteration ($Q=50$) and a total of 2000 iterations are performed for each encoder ($I=2000$). A comparison of the reconstruction loss is plotted in Fig. \hyperref[fig:cifar-recon]{3(a)}.
In this experiment, images of size 3x32x32 (3072 features) were compressed to a size of 16x4x4 (256 features) thereby achieving a factor of compression of more than 12. We note that the proposed method achieves a lower reconstruction error when compared to a single AE while maintaining the same factor of compression. In addition, we also note that the convergence of loss is much faster as compared to a single AE. 

\subsubsection{Experiments on the Fashion MNIST Dataset} 
\begin{figure}[t]
\centering 
\subfigure[CIFAR-10 dataset]{
  \includegraphics[scale=0.5]{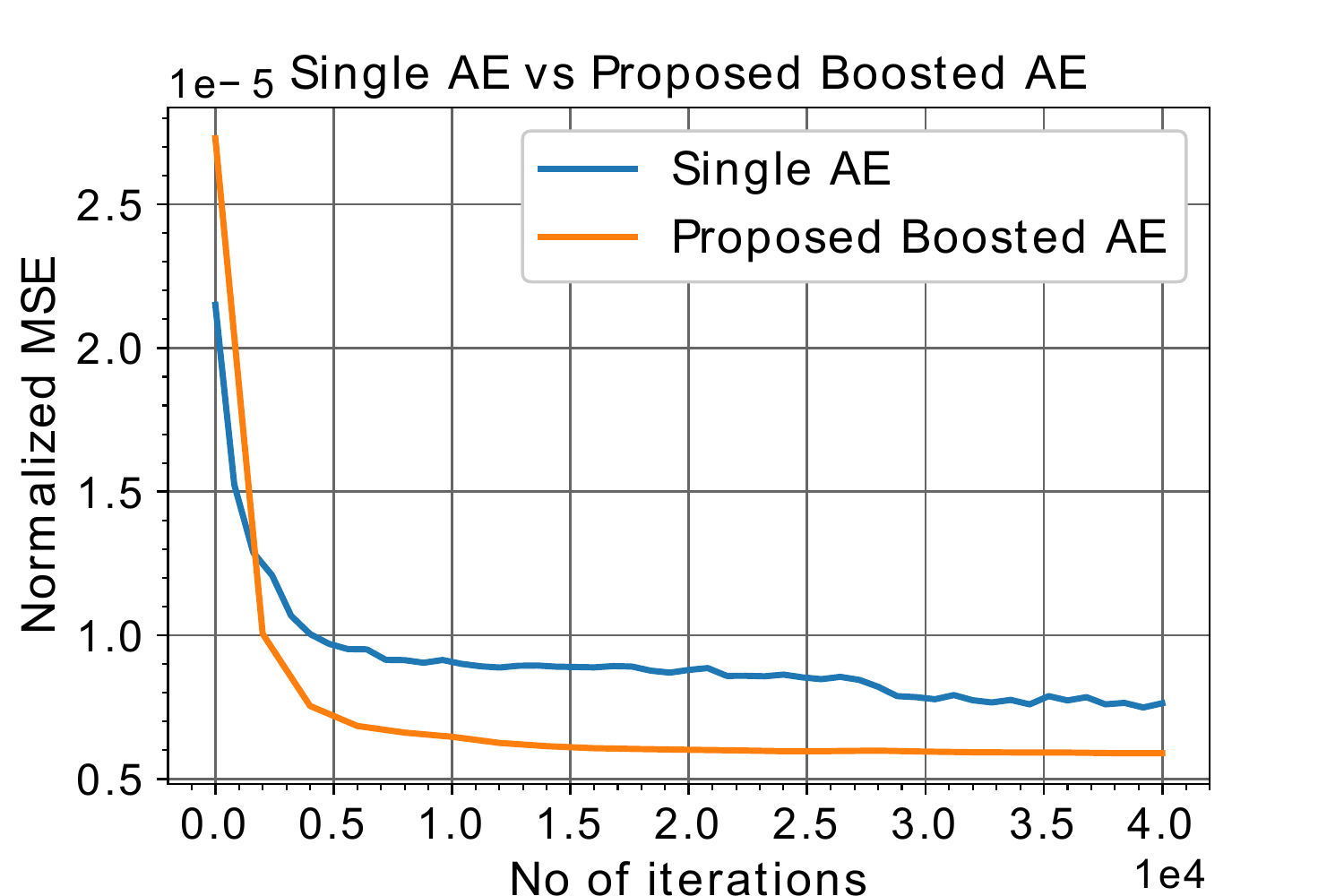}
  \label{fig:cifar-recon}}
\hspace*{-2.9em}
\subfigure[F-MNIST dataset]{
  \includegraphics[scale=0.5]{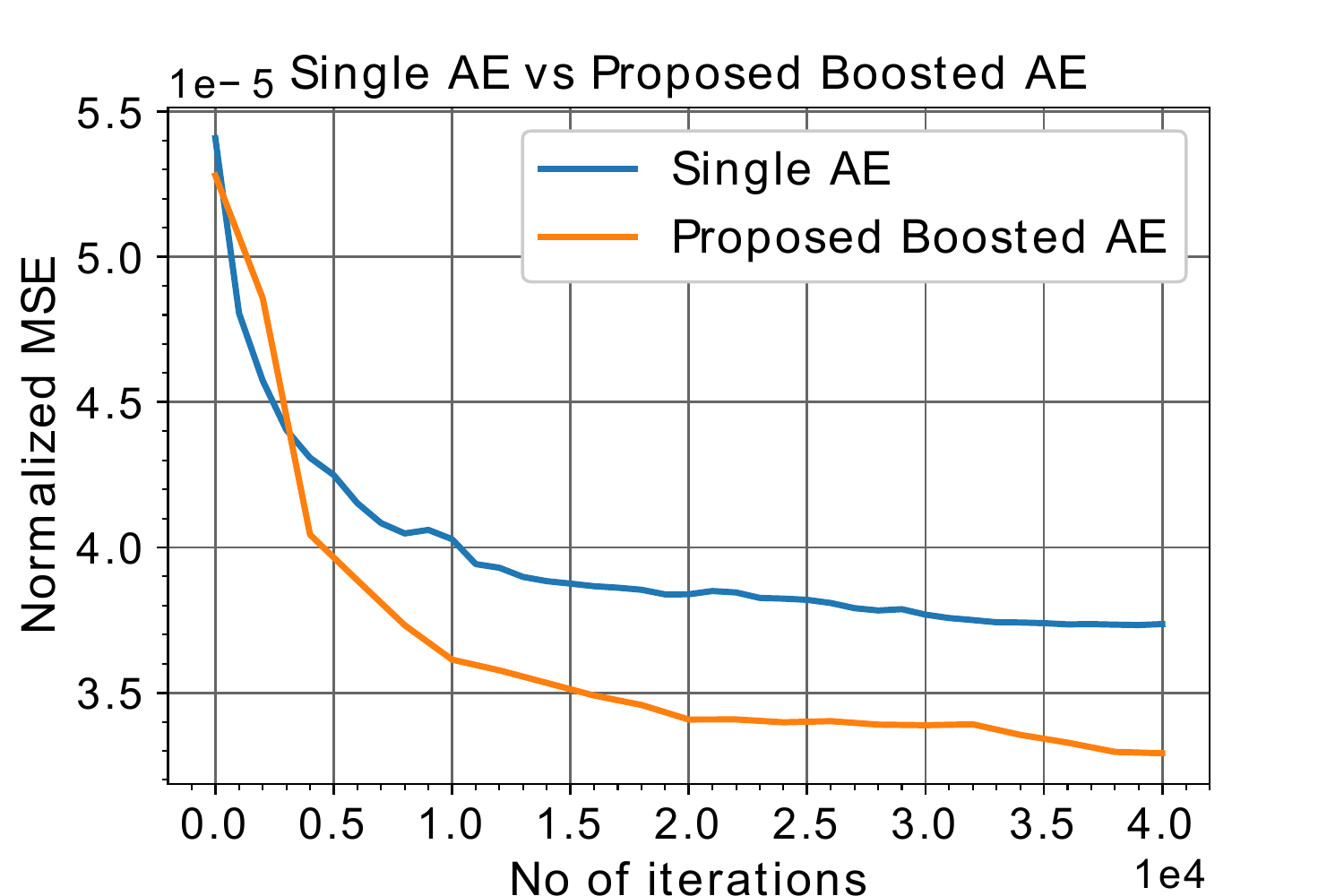}
  \label{fig:fmnist-recon}
\label{fig:recon}}
\caption{Reconstruction error during validation}
\end{figure}

The F-MNIST dataset contains 70,000 32x32 grey scale images in 10 different classes of which 50000 images were used for training, 10000 for validation and 10000 images for testing. The architecture of the encoder used is Conv2D(1,2,4,2,1)-Conv2D(2,4,4,2,1)-Conv2D(4,8,3,2,1)-Conv2D(8,8,4,2,1). The decoder is constructed symmetrically to that of the encoder. The activation functions are the same as in the CIFAR-10 experiment.  The Adam optimizer was used in training both the single and the ensemble of AEs with a learning rate of 5e-3. A single AE was trained for 40 epochs using a batch size of 50. For the proposed boosted AE, we consider $M=20$; 50 data samples are chosen for each iteration ($Q=50$) and a total of 2000 iterations are performed for each encoder ($I=2000$). The reconstruction error for a single AE as well as the proposed method is plotted in Fig. \hyperref[fig:fmnist-recon]{3(b)}.

 In the Fashion MNIST experiment, images of size 1x28x28 (784 features) were compressed to a size of 8x2x2 (32 features). The images have been compressed by more than a factor of 24 and once again, the proposed ensemble method outperforms a single AE in terms of both reconstruction error as well as  convergence. 
 
\section{Applications of Boosted Autoencoders}
AEs are used for varied applications such as anomaly detection, clustering (as a dimension reduction technique), de-noising of images, image classification, generative applications, etc. The applications can be broadly grouped based on whether the  reconstructed output or the encoded output is used. In this section, we employ the proposed boosting method to an application from each of these groups to demonstrate its utility. 

\subsection{Anomaly detection}
Anomaly detection refers to the task of finding unusual instances that stand out from the normal data. AEs have been widely employed in detecting anomalies in both images and videos (\cite{gong2019memorizing},\cite{hasan2016learning}). The concept behind using AEs for anomaly detection is as follows: when an AE trained on normal samples encounters an anomaly, it results in a high reconstruction error. For our experiment, a semi-supervised learning setting is considered where all the data points in the training set are normal, whereas the testing data contains both normal and anomalous samples (Refer to \cite{villa2021semi} for a thorough survey on semi-supervised learning for anomaly detection).

\subsubsection{Simulation setting} 
We conduct our experiments on two well-known datasets: CIFAR-10, and Fashion-MNIST (F-MNIST). Following the setting in \cite{gong2019memorizing}, \cite{zong2018deep} and \cite{zhai2016deep}, 10 anomaly detection (i.e. one-class classification) datasets are constructed by sampling images from each class as normal samples and sampling anomalies from the remaining classes. The training data only consists of normal samples, and 10$\%$ of the training data is used for validation. The test data contains equal number of samples from all the 10 classes.

Following \cite{ruff2018deep}, we use CNNs similar to LeNet, i.e., each Convolutional layer is followed by a 2x2 Max Pooling layer with stride=2 and a leaky ReLU activation layer. The leakiness of ReLU activations is set to be $\alpha$ = 0.1. For CIFAR-10, the encoder has 3 convolutional layers: Conv2D(3,32,3,1,1)-Conv2D(32,64,3,1,1)-Conv2D(64,64,3,1,1), followed by a dense layer of 256 units. For F-MNIST, we used a fully connected encoder. This shows that the proposed method is not limited to convolutional encoders. The encoder had 4 dense layers: 512-256-128-50. In all the cases, the decoder was constructed symmetrically to the encoder, replacing Max-Pooling with Upsampling. 

The Adam optimizer (\cite{kingma2014adam}) was used in training both the single and the ensemble of AEs with a learning rate of 3e-3. The single AE was trained for 100 epochs using a batch size of 50. For the proposed boosted AE, we consider an ensemble of 5 networks ($M=5$); 50 data samples are chosen for each iteration ($Q=50$). For CIFAR-10 a total of 1800 iterations are performed for each encoder ($I=1800$), while for F-MNIST, a total of 2000 iterations are performed for each encoder ($I=2000$).

\subsubsection{Baselines}
The proposed method is compared with the following existing methods for anomaly detection:
\begin{itemize}
    \item One Class SVM (OC-SVM/SVDD) (\cite{scholkopf1999support}): This algorithm captures the density of the data and classifies examples on the extremes of the density function as anomaly. \taddRev{The hyperparameter selection is as done in \cite{ruff2018deep}.}
    \item Isolation Forest (IF) (\cite{liu2008isolation}): This algorithm ‘isolates’ observations by randomly selecting a feature and then randomly selecting a split value between the maximum and minimum values of the selected feature. This random partitioning produces noticeable shorter paths for anomalies. We set the number of trees = 100 and sub\_sample size = 256.
    \item Kernel Density Estimation (KDE): This algorithm uses multiple Gaussian Kernels to estimate the density of the distribution. We chose the bandwidth of the Gaussian Kernels from \{2$^{0.5}$,2$^{1}$,...,2$^{5}$\} using the log-likelihood score and a 5-fold Cross Validation.
    \item Bagging Random Miner (BRM) (\cite{camina2019bagging}): This algorithm uses an ensemble of random miners; each random miner randomly chooses data from the training set and then creates a set of most representative objects (MROs). Then a covariance matrix is computed over these MROs, which is then used to calculate the average object pair distance (AOPD). This AOPD will then be used as a threshold to classify the outliers. \cite{villa2021semi} has named it as the best semi-supervised classifier for anomaly detection.
    \item One Class K-Means with with Randomly-projected features Algorithm (OCKRA) (\cite{rodriguez2016ensemble}): This algorithm uses an ensemble of one class K-means, each trained on a subset of features, which are randomly selected and is named the second best semi-supervised classifier for anomaly detection by \cite{villa2021semi}. The hyperparameters are tuned according to \cite{rodriguez2016ensemble}.
\end{itemize}
In all the above methods, the data is compressed using PCA while maintaining 95\% of the variance \taddRev{whereas our method does not employ any pre-processing technique.}
\begin{table}[H]
    \centering
    \begin{tabular}{| l | l | l | l | l | l|l|l|}
\hline Class & OC-SVM & IF & KDE & OCKRA&BRM& DAE & Boosted AE
\\ \hline Airplane & 0.6123 & 0.5186 & 0.6298 &0.5594&0.5965& 0.6268 & \textbf{0.6657} 
\\ \hline Automobile & 0.5097& 0.5016 & 0.5033&0.5067 &0.5001& 0.5012&\textbf{0.5144}
\\ \hline Bird & 0.6041 & 0.5213 & 0.6276 &0.5911&0.6212& 0.6249 &\textbf{0.6603}
\\ \hline Cat & 0.5086 & 0.5026 & 0.5125 &0.4985&0.5273& 0.5359 &\textbf{0.5597}
\\ \hline Deer & \textbf{0.6903} & 0.5389 & 0.6769& 0.6542&0.6423& 0.6702&0.6838
\\ \hline Dog & 0.5205 & 0.5035 & 0.5317&0.5083&0.5249& \textbf{0.5386}&0.5334
\\ \hline Frog & 0.6488 & 0.5167 & 0.6788 &0.6543&0.5938& 0.6709&\textbf{0.6982}
\\ \hline Horse & 0.5202 & 0.503 & 0.5307 &0.5232&0.5032& \textbf{0.5355}&0.5280
\\ \hline Ship & 0.6150 & 0.5398& 0.6239 &0.6214&0.6246& 0.6635&\textbf{0.6734}
\\ \hline Truck & \textbf{0.5381} & 0.5554 & 0.4742 &0.5206&0.4986&0.5043 &0.4996
\\ \hline
\end{tabular}
\caption{CIFAR-10 Classwise AUC}
    \label{tab:ano_cifar}
\end{table}

\begin{table}[H]
    \centering
    \begin{tabular}{| l | l | l | l | l | l| l| l |}
\hline Class & OC-SVM & IF & KDE & OCKRA & BRM & DAE &  Boosted AE 
\\ \hline T-Shirt & 0.8303 & 0.6932 & 0.8343 &0.8243& 0.8354& 0.8405 & \textbf{0.8425} 
\\ \hline Trouser & 0.8978& 0.8982 & 0.9345 &0.9141&0.9380& 0.9529 & \textbf{0.9546} 
\\ \hline Pullover & 0.8150 & 0.6851 & 0.8104&0.8078 &0.8057& 0.8134 &\textbf{0.8253} 
\\ \hline Dress & 0.8525 & 0.7653 & 0.8631&0.8468 &0.8528& 0.8609 &\textbf{0.8858}
\\ \hline Coat & 0.8235 & 0.7228 & 0.8303 &0.8142&0.8230& \textbf{0.8581} &0.8344
\\ \hline Sandals & 0.7831 & 0.7711 & \textbf{0.8162} &0.7865&0.8099& 0.7868 &0.7933
\\ \hline Shirt & \textbf{0.7706 }& 0.6791 & 0.7359 &0.7491&0.7550& 0.7395 &0.7488
\\ \hline Sneaker & 0.9253 & 0.8823 & 0.9098 &0.9148&0.9004& 0.9237&\textbf{0.9311}
\\ \hline Bag & 0.7666 & 0.5997 & 0.7716&0.7675&0.7579&0.7700 &\textbf{0.7928}
\\ \hline Ankle Boots & 0.9056 & 0.8263 &0.8787&0.8933&0.8914& 0.8827 & \textbf{0.9064}
\\ \hline
\end{tabular}
\caption{F-MNIST Classwise AUC}
    \label{tab:ano_fmnist}
\end{table}


 We use the Area Under Receiver Operation Characteristic curve (AUC-ROC) as a metric to quantify the efficiency of anomaly detection (\cite{ruff2018deep},\cite{abati2019latent}, \cite{gong2019memorizing}). An ROC curve plots True Positive Rate (TPR) vs. False Positive Rate (FPR) at different classification thresholds. AUC-ROC provides an aggregate measure of performance across all possible classification thresholds. The class-wise AUC-ROC values are tabulated for CIFAR-10, and F-MNIST in Tables \ref{tab:ano_cifar}, and \ref{tab:ano_fmnist} respectively.
 
Our proposed ensemble method consisting of 5 boosted encoders (termed 'Boosted AE') is compared to the methods listed above and the best performing method in each class is highlighted in bold. For CIFAR-10 and F-MNIST datasets, we note that the proposed method outperforms the others in most of the classes and is only slightly lower that the best method in other classes. This illustrates that the proposed boosting method is efficient while the application uses the reconstructed output. 

\subsection{Clustering}
Clustering is the task of dividing unlabelled data points into groups based on their similarity. Traditional clustering  algorithms such  as \cite{macqueen1967some} and \cite{ng2001spectral} classify input data into the same class based on the similarity of extracted features. These methods cannot be applied directly to image data due to their high dimensions (\cite{zhu2020image}). An autoencoder allows the user to represent a high dimensional image in a lower-dimensional space and has recently become a popular choice as a pre-processing step for clustering \cite{song2013auto}. 

We use a single AE and the proposed boosted AE as a pre-processing step for clustering and compare them with a traditional dimensionality reduction technique, Principal Component Analysis (PCA). After dimensionality reduction, we use the K-means proposed by \cite{lloyd1982least} algorithm for clustering. \taddRev{Note that any clustering algorithm may be used, we use K-means as an example to demonstrate the efficiency of the proposed method as a pre-processing technique.}  Normalized Mutual Information (NMI) is used as our evaluation metric; NMI is a normalization of the Mutual Information (MI)  score \taddRev{within the grouped classes} to scale the results between 0 (no mutual information) and 1 (perfect correlation). It is computed as  $$\frac{2*I(Y;C)}{H(Y)+H(C)}$$
where $I$, $H$, $Y$ and $C$ refer to Mutual Information, entropy, class labels and cluster labels respectively. 

The network architecture used is once again similar to LeNet, i.e., each convolutional layer is followed by a 2x2 Max Pooling layer with stride=2 and a leaky ReLU activation layer. The leakiness of ReLU activations is set to be $\alpha$ = 0.1. For both F-MNIST and MNIST, the encoder has 2 convolutional layers: Conv2D(1,8,5,1,0)-Conv2D(8,4,5,1,0), followed by a dense layer of 10 units. The experiments are performed in MNIST and F-MNIST datasets and the results obtained are tabulated in Table \ref{tab:clustering}.


\begin{table}[H]
    \centering
    \begin{tabular}{| l | l | l | l | l |}
\hline Method & MNIST & F-MNIST 
\\ \hline PCA+K-means & 0.4800 &  0.5005
\\ \hline Single AE+K-means & 0.6476 & 0.5331 
\\ \hline Ensemble Method+K-means & \textbf{0.6900} & \textbf{0.5687} 
\\ \hline
\end{tabular}
\caption{NMI scores for clustering}
    \label{tab:clustering}
\end{table}
It is observed from Table \ref{tab:clustering} that using AEs for reducing the dimension of the data is more efficient that PCA. It is also observed that boosting AEs lead to an improvement in the NMI score when compared to the use of a single AE. \taddRev{Our results demonstrate the improved performance of the proposed boosted AE as a pre-processing technique for clustering.} This shows that the proposed method helps in enhancing the performance of an AE when the encoded representation is used as well.

\section{Conclusion and Future Work}
In this work, we introduced a boosted ensemble of encoders \taddRev{as an effective way of boosting AEs for applications that use either reconstructed or encoded outputs.} Through various experiments, we have shown that our method performs significantly better than a single AE. Our method can also be extended to works that propose modifications to enhance the performance of a single AE. For example, \cite{zhu2020image} uses AE for clustering, but incorporates Predefined Evenly-Distributed Class Centroids and MMD Distance in their loss function. The modification of a single AE can be combined with our proposed boosting architecture for a potential improvement in the performance. Similarly, \cite{gong2019memorizing} and \cite{ruff2018deep} have used modifications of AEs for anomaly detection. These methods can also be combined with our proposed boosting ensemble. We believe that our method provides the first step towards a universal boosting framework for AEs. It can be further improved by exploring a variety of modifications such as incorporating a weighted average of the encoded outputs, employing heterogeneous networks in the ensemble (for instance, different depths/widths/activations), etc.

\bibliography{main}






\end{document}

%% file: autoencoder_schematic.tex
\tikzset
{
  myTrapezium/.pic =
  {
    \draw [fill=blue!50] (0,0) -- (0,\b) -- (\a,\c) -- (\a,-\c) -- (0,-\b) -- cycle ;
    \draw [dashed] (-0.2,\c+0.1) -- (-0.2,-\c-0.1);
    \draw [dashed] (\a+0.2,\c+0.1) -- (\a+0.2,-\c-0.1);
    \draw [dashed] (\a+0.2,\c+0.1) -- (-0.2,\c+0.1);
    \draw [dashed] (\a+0.2,-\c-0.1) -- (-0.2,-\c-0.1);
    \coordinate (-center) at (\a/2,0);
    \coordinate (-out) at (\a,0);
  },
  myArrows/.style=
  {
    line width=1mm, 
    black,
    -{Triangle[length=1.5mm,width=2mm]},
    shorten >=1pt, 
    shorten <=1pt, 
  }
}
    \def\a{2}  
    \def\b{1} 
    \def\c{2}  
\begin{figure}[H]
\centering
\begin{tikzpicture}
[
  node distance=10mm, 
  every node/.style={align=center},
]

  \node (middleThing) 
  {\begin{tabular}{r}\end{tabular}};
  \pic (right)[right=of middleThing.east] {myTrapezium} ;
  \pic (left)[left=of middleThing.west, rotate=180] {myTrapezium} ;
  \node at (left-center) {\small{Encoder}} ;
  \node at (right-center) {\small{Decoder}} ;
  \node {\LARGE{$\textbf{h}$}} ;

  \def\d{0.9}
  \coordinate (u) at (\d,0);
  \draw [myArrows] ($(left-out)+2.3*(u)$) -- ++(u) node [anchor=west]{};
  \draw [myArrows] ($(right-out)-3.3*(u)$) -- ++(u) node [anchor=west]{};
  \draw [myArrows] (right-out) -- ++(u) node [anchor=west]
  {\LARGE{$\textbf{y}$}};
  \draw [myArrows] ($(left-out)-(u)$) node [anchor=east] {\LARGE{$\textbf{x}$}} -- ++(u) ;
  
\end{tikzpicture}

    \caption{Schematic of an autoencoder}
    \label{fig:schematic}
\end{figure}

%% file: ensemble_schematic.tex
\begin{figure}[h]
    \centering

\tikzset
{
  myTrapezium/.pic =
  {
    \draw [fill=blue!50] (0,0) -- (0,\b) -- (\a,\c) -- (\a,-\c) -- (0,-\b) -- cycle ;
    \draw [dashed] (-0.2,\c+0.1) -- (-0.2,-\c-0.1);
    \draw [dashed] (\a+0.2,\c+0.1) -- (\a+0.2,-\c-0.1);
    \draw [dashed] (\a+0.2,\c+0.1) -- (-0.2,\c+0.1);
    \draw [dashed] (\a+0.2,-\c-0.1) -- (-0.2,-\c-0.1);
    \coordinate (-center) at (\a/2,0);
    \coordinate (-out) at (\a,0);
  },
  myensemble/.pic =
  {
    \draw [fill=blue!50] (0,-1.6) -- (0,-1.6+\b/5) -- (\a,-1.6+\c/5) -- (\a,-1.6-\c/5) -- (0,-1.6-\b/5) -- cycle ;
    \node at (0.1, -1.6) {\small{Encoder 1}} ;
    \draw [fill=blue!50] (0,-0.6) -- (0,-0.6+\b/5) -- (\a,-0.6+\c/5) -- (\a,-0.6-\c/5) -- (0,-0.6-\b/5) -- cycle ;
    \node at (0.1, -0.6) {\small{Encoder 2}} ;
    \draw [fill=blue!50] (0,1.6) -- (0,1.6+\b/5) -- (\a,1.6+\c/5) -- (\a,1.6-\c/5) -- (0,1.6-\b/5) -- cycle ;
    \node at (0.1, 1.6) {\small{Encoder M}} ;

    \draw [dotted] (1.1,1.1) -- (1.1,-0.1);
    \draw [dashed] (-0.2,\c+0.1) -- (-0.2,-\c-0.1);
    \draw [dashed] (\a+0.2,\c+0.1) -- (\a+0.2,-\c-0.1);
    \draw [dashed] (\a+0.2,\c+0.1) -- (-0.2,\c+0.1);
    \draw [dashed] (\a+0.2,-\c-0.1) -- (-0.2,-\c-0.1);
    \coordinate (-center) at (\a/2,0);
    \coordinate (-out) at (\a,0);
  },
  myArrows/.style=
  {
    line width=1mm, 
    black,
    -{Triangle[length=1.5mm,width=2mm]},
    shorten >=1pt, 
    shorten <=1pt, 
  }
}
    \def\a{2}  
    \def\b{1} 
    \def\c{2}  
\begin{tikzpicture}
[
  node distance=10mm, 
  every node/.style={align=center},
]

  \node (middleThing) 
  {\begin{tabular}{r}\end{tabular}};
  \pic (right)[right=of middleThing.east] {myTrapezium} ;
  \pic (left)[left=of middleThing.west, rotate=180] {myensemble} ;
  \node at (right-center) {\small{Decoder}} ;
  \node {\LARGE{$\textbf{h}$}} ;

  \def\d{0.9}
  \coordinate (u) at (\d,0);
  \draw [myArrows] ($(left-out)+2.2*(u)$) -- ++(u) node [anchor=west]{};
  \draw [myArrows] ($(right-out)-3.2*(u)$) -- ++(u) node [anchor=west]{};
  \draw [myArrows] (right-out) -- ++(u) node [anchor=west]
  {\LARGE{$\textbf{y}$}};
  \draw [myArrows] ($(left-out)-(u)$) node [anchor=east] {\LARGE{$\textbf{x}$}} -- ++(u) ;
  
\end{tikzpicture}
 \caption{Proposed architecture for boosting autoencoders }
    \label{fig:ensemble}
\end{figure}